\documentclass[letterpaper, 10 pt, conference]{ieeeconf}  

\IEEEoverridecommandlockouts                              

\usepackage{graphics} 
\usepackage{epsfig} 
\usepackage{mathptmx} 
\usepackage{times} 
\usepackage{amsmath} 
\usepackage{amssymb}  
\usepackage{multirow}
\usepackage{booktabs}
\usepackage{tabularx}
\usepackage{cite}
\usepackage{xcolor}
\usepackage{titlesec}
\usepackage{subcaption} 
\usepackage{algorithm,algpseudocode}
\usepackage{dblfloatfix}
\usepackage{comment}
\usepackage{bm}

\newcommand{\vista}{\textit{\textbf{VISTA}}}
\newcommand{\vistawithversion}{\textit{\textbf{VISTA 2.0}}}
\newcommand{\ILplain}{IL}

\newcommand{\GPLplain}{GPL}

\newcommand{\figref}{Fig.~\ref}
\newcommand{\tabref}{Tab.~\ref}
\newcommand{\secref}{Sec.~\ref}
\newcommand{\norm}[1]{\left\lVert#1\right\rVert}

\renewcommand{\baselinestretch}{0.965}

\title{\LARGE \bf
\textit{VISTA 2.0}: An Open, Data-driven Simulator for Multimodal Sensing and Policy Learning for Autonomous Vehicles
}

\author{Alexander Amini$^{1,*}$, Tsun-Hsuan Wang$^{1,*}$, Igor Gilitschenski$^{2}$, Wilko Schwarting$^{1}$, Zhijian Liu$^{1}$,\\Song Han$^{1}$, Sertac Karaman$^{1}$, and Daniela Rus$^{1}$
\thanks{* The first two authors have contributed equally to this work. This work was supported by National Science Foundation and Toyota Research Institute. We gratefully acknowledge the support of NVIDIA with the donation of the Drive AGX Pegasus.}
\thanks{$^{1}$ Department of Electrical Engineering and Computer Science (EECS), Massachusetts Institute of Technology (MIT). {\tt\{amini,tsunw,wilkos,zhijian,song,sertac,rus\}@mit.edu}}%
\thanks{$^{2}$ Department of Computer Science, University of Toronto and Toyota Research Institute (TRI). {\tt gilitschenski@cs.utoronto.edu}}%
}

\begin{document}

\maketitle
\thispagestyle{empty}
\pagestyle{empty}

\begin{abstract}

Simulation has the potential to transform the development of robust algorithms for mobile agents deployed in safety-critical scenarios. 
However, the poor photorealism and lack of diverse sensor modalities of existing simulation engines remain key hurdles towards realizing this potential. 
Here, we present \vista{}, an open source, data-driven simulator that integrates multiple types of sensors for autonomous vehicles. 
Using high fidelity, real-world datasets, \vista{} represents and simulates RGB cameras, 3D LiDAR, and event-based cameras, enabling the rapid generation of novel viewpoints in simulation and thereby enriching the data available for policy learning with corner cases that are difficult to capture in the physical world. 
Using \vista{}, we demonstrate the ability to train and test perception-to-control policies across each of the sensor types and showcase the power of this approach via  deployment on a full scale autonomous vehicle. 
The policies learned in \vista{} exhibit sim-to-real transfer without modification and  greater robustness than those trained exclusively on real-world data.

\end{abstract}

\section{Introduction}

Simulation has emerged as an essential tool for advancing new algorithms in robot perception, learning, and evaluation~\cite{dosovitskiy2017carla,Deitke2020,shah2017airsim}. For safety-critical domains in particular, such as for autonomous vehicles, experience in simulation is often significantly faster and safer than direct operation in the physical world. Simulation affords the potential to rapidly synthesize novel data for training, including challenging edge cases difficult to capture in the real world~\cite{dosovitskiy2017carla, Deitke2020}. An agent's exposure to edge cases during training is critical to achieving robustness to out-of-distribution events. Furthermore, high-fidelity, in-simulation testing could improve an agent's performance when deployed into safety-critical, human-centric environments. Thus, simulation could enable the development of algorithms and models better equipped to handle the diverse challenges of the physical world, facilitating their deployment on embodied mobile agents.

\begin{figure}[!t]
\centering
\vspace{-3pt}
\includegraphics[width=0.85\linewidth]{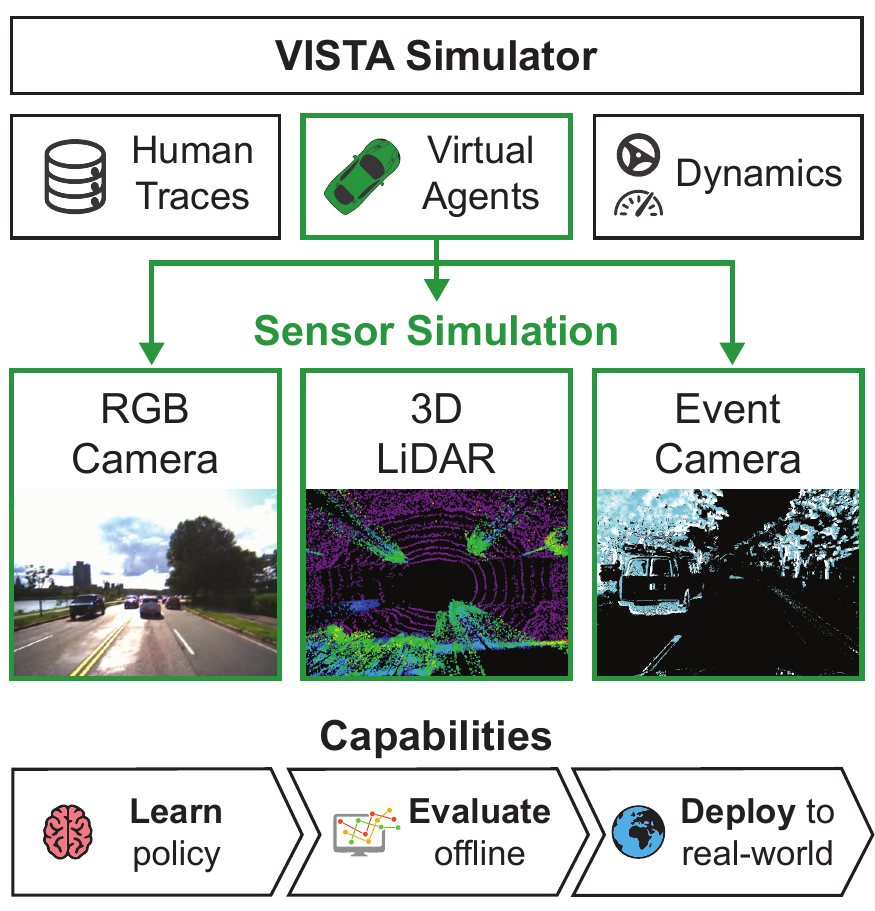}
\caption{\textbf{\vistawithversion} is an open-source data-driven simulator for multi-sensor perception of embodied agents. Leveraging data of the real-world, \vista{} synthesizes ego-agent viewpoints as their dynamics unroll novel trajectories in the environment. Sensors are efficient and high fidelity for online perception learning, evaluation, and sim-to-real deployment.}
\label{fig:teaser}
\vspace{-20pt}
\end{figure}

Despite the potential of simulation, the stark lack of photorealism and a paucity of diverse simulated sensor representations have remained crucial barriers towards realizing this promise. Data-driven simulation, unlike traditional model-based simulation, synthesizes novel viewpoints directly from real data and has emerged as an approach to overcome the photorealism and sim-to-real gap~\cite{amini2020learning}. However, issues in scaling simulation engines to multiple sensor types for online perception learning remain. Because embodied agents benefit from rich perception~\cite{xiao2019multimodal}, integrating multiple sensor modalities could facilitate adaptation to a wide variety of environmental conditions (e.g., combining LiDAR and camera feedback to stay on the road in low visibility lighting). There remains a need for unified, flexible, and open-source data-driven simulation engines to fuel the development of new algorithms for embodied agent learning and evaluation.

In this paper, we present \vistawithversion{}, a multi-sensor, data-driven engine for autonomous vehicle simulation, perception, and learning (Fig.~\ref{fig:teaser}). \vista{} synthesizes novel viewpoints consistent with each sensor representation, simulates agents in synthesized scenarios, and supports large scale learning and testing environments. Our work prioritizes a lightweight API for processing existing real-world datasets, operating only on local viewpoint changes to achieve efficient computational rendering and low memory costs.

We develop novel view synthesis capabilities for three distinct sensors: 2D RGB cameras, 3D LiDARs, and asynchronous event-based sensors. Using \vista{} to generate data, we train end-to-end (i.e., perception-to-control) policies using using guided policy learning (GPL)\cite{levine2013guided} and demonstrate direct policy transfer onto a full-scale autonomous vehicle. Further, our results highlight the importance of simulation in two key contexts. First, we show that simulating novel viewpoints can drastically improve the robustness of learned policies and a vehicle's ability to recover from challenging off-orientation positions. Second, for sensors which conflate the ego-motion of the agent with the control decision (e.g., event-based cameras), we find that existing state-of-the-art imitation learning (IL) approaches~\cite{maqueda2018event} cannot achieve closed-loop success using only real-world data. Using \vista{}, we not only uncover this issue, but also overcome it by decoupling the learning signal within \vista{} to successfully train closed-loop controllers from event-based sensors.

In summary, the contributions of this paper are as follows:
\begin{enumerate}
    \item \vistawithversion{}, an open-source, multi-sensor, data-driven simulator for learning and evaluating autonomous vehicle perception and control.
    \item A framework for translating real-world data to a simulated perception-control API spanning a diversity of compatible environments with varying complexity, lighting, weather, and road types.
    \item End-to-end autonomous vehicle control policies using each sensor type, learned within \vista{} and directly deployed on a full-scale vehicle. Learned policies exhibit direct sim-to-real transfer and improved robustness than those trained solely on real-world data.
\end{enumerate}

\section{Related Work}

\textbf{Cross-Sensor Transfer:} Numerous works consider augmenting sensing modalities and simulating different modalities via, often learned, sensor fusion.
For example, in \textit{Monocular Depth Prediction} approaches, a neural network is typically trained to act as a depth sensor using monocular image data~\cite{godard2017unsupervised,Fu2018}.
Similarly, \textit{Depth Completion} combines cameras with sparse depth from LiDAR to simulate a dense depth sensor~\cite{Xu2019,Zhao2021}.
In event-based vision, recent work combine the use of classical and event-based cameras for simulating higher frame-rate cameras~\cite{Tulyakov2021}, depth predictions~\cite{HidalgoCarrio2020,Gehrig2021}, or focus on translating between these two modalities~\cite{rebecq2018esim,Rebecq2019,Gehrig2020,ParedesValles2021}. 
We draw inspiration from these works but focus on a unified simulation framework that jointly simulates a diverse set of sensors while supporting novel view synthesis.

\textbf{Simulation:} 
The use of simulation for learning and robotics has exploded in recent years. Model-driven simulators rely on predefined models of scenery and underlying physics~\cite{Coumans2021,Tassa2020,Todorov2012, drake, wymann2000torcs}. Several model-based engines that focus on high quality visual appearances are widely used for autonomous vehicles~\cite{dosovitskiy2017carla, muller2018sim4cv} and drones~\cite{Shah2017, song2020flightmare}. These engines rely on heavily engineered video-game rendering platforms, but still lack the photorealism necessary for direct policy transferability. In contrast, data-driven simulators~\cite{Xia2018, Savva2019, amini2020learning, Deitke2020, manivasagam2020lidarsim, Wang2020} present greater photorealism by leveraging real data to reconstruct virtual worlds of the scene before synthesizing novel views. Our work follows this line of research with a focus on local scene synthesis for scalability and on learning transferable policies for embodied AI research.

\textbf{Driving Policy Learning:} While policy learning for driving using real-world data is largely restricted to IL~\cite{bojarski2016end, xu2017end, amini2019variational, lechner2020neural, hawke2020urban},  learning in simulation allows for greater algorithmic flexibiltiy ranging from IL~\cite{rhinehart2018deep, xiao2019multimodal, codevilla2018end}, to RL~\cite{pan2017virtual, dosovitskiy2017carla, amini2020learning, fuchs2021super}, and GPL~\cite{levine2013guided, chen2020learning}. Evaluation of trained policies in closed-loop simulation~\cite{bojarski2016end, codevilla2018offline, amini2018learning, rhinehart2018deep, xiao2019multimodal, chen2020learning} also presents benefits over open-loop evaluation~\cite{xu2017end, amini2018variational, maqueda2018event}. Similarly, our work leverages simulation for edge-case training data generation, and closed-loop evaluation before deployment.



\section{Multi-sensor Simulation}
\begin{figure*}[!t]
\centering
\includegraphics[width=\linewidth]{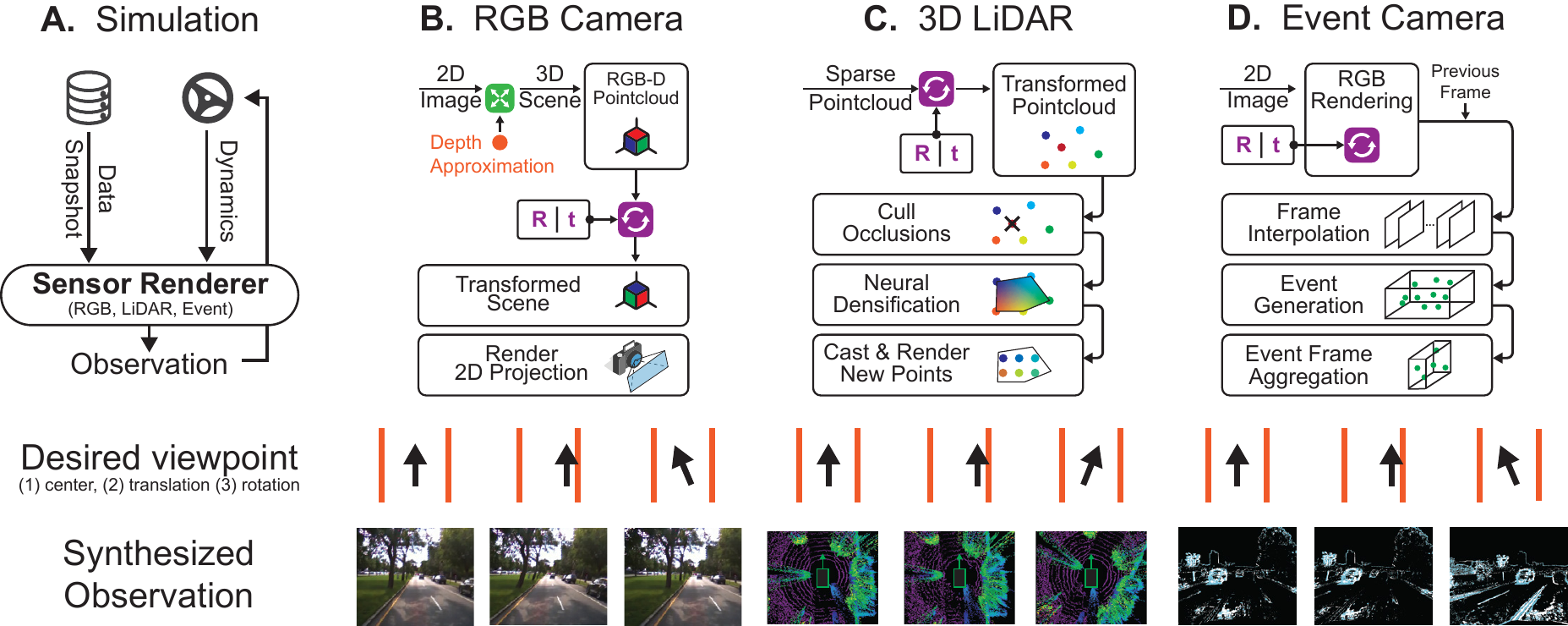}
\caption{\textbf{Multi-sensor simulation.} (A) Virtual agent dynamics unroll trajectories in a data-driven environment, at each time rendering a sensor observation from the novel viewpoint at that time. Three types of perception sensors are able to be synthesized including images from RGB cameras (B), 3D pointclouds from LiDAR (C), and continuous differential events (D). Examples of novel viewpoint synthesis are visualized for each sensor. }
\label{fig:overview}
\vspace{-15pt}
\end{figure*}

\subsection{Background}
\vista{} \textit{\textbf{1.0}} \cite{amini2020learning}~is a data-driven simulator that synthesizes RGB images at novel viewpoints around the local trajectory of a dataset. The precollected image sequence represents a sparsely sampled representation of a continuous trajectory traversed by a vehicle in the physical world. Any novel viewpoints can be associated with a frame with the closest pose and rerendered to the virtual agent's position. The overall pipeline of \vista~is: (1) update vehicle state with a continuous kinematic model, (2) retrieve the closest frame in the dataset with respect to the current pose, (3) project the frame into 3D space to reconstruct the scene, and (4) reproject back into the ego-agent's point-of-view. Please refer to~\cite{amini2020learning} for details. The goal of \vistawithversion~is to extend simulation to other modalities in a data-driven manner, namely synthesizing novel sensor measurements of LiDAR and event data locally around the dataset, and to leverage and release this platform for robust perception learning.


\subsection{LiDAR synthesis}

LiDAR sensors play a central role in modern autonomy pipelines due to their accuracy in measuring geometric depth information and robustness to environmental changes like illumination. Unlike cameras which return structured grid-like images, the LiDAR sensor captures a sparse pointcloud of the environment. Here, every point is represented by a 4-tuple: $(x, y, z, i)$, where $(x,y,z)$ is the position of the point in 3D Cartesian space and $i$ is the intensity feature measurement of that point. Given a virtual agent's position in the environment, along with a relative transformation (rotation $R\in\mathbb{R}^{3\times 3}$, and translation $t\in\mathbb{R}^3$) to the nearest human collected pointcloud, $\Psi$, the goal of \vista{} is to synthesize a novel LiDAR pointcloud, $\Psi'$, which appears to originate from the virtual agent's relative position. 

\begin{figure}[!b]
\centering
\includegraphics[width=\linewidth]{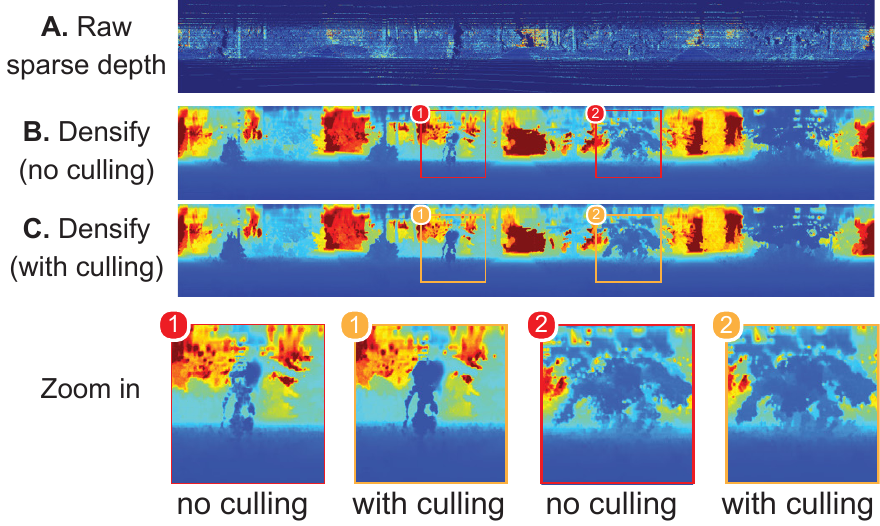}
\caption{\textbf{Culling occluded points.} Transformed sparse scenes (A) will have points which should be rejected (culled) before rendering to avoid blending of foreground and background (B). Our culling algorithm (C) is lightweight, GPU-accelerated, and does not rely on raycasting a scene mesh. }
\label{fig:lidar_culling}
\end{figure}

Since $\Psi$ is represented in 3D Cartesian space, a naive solution would be to directly apply the relative transformation of the agent $(R, t)$ to $\Psi$ as a rigid transformation: $\Psi' = R\, \Psi + t$. However, this approach will fail for several reasons. The pointcloud obtained from a LiDAR sensor has a specific ring pattern pattern originating at the sensor's optical center. Applying a rigid transformation to the points will not only transform the individual points, but in doing so, also transform and break the ring structure inherently defining the sensor's location. Instead, to preserve the sensor structure we must recast LiDAR rays, from the new sensor location, into the scene and estimate new readings. Furthermore, the naively transformed pointcloud will very likely have points which may have been visible in the original scan, but become occluded in the new viewpoint and thus need to be rejected to maintain line-of-sight properties of the sensor. To overcome these issues, we (1) cull the now occluded points, (2) create a dense representation of the sparse pointcloud, and (3) sample from the dense representation according to a sensor-specific prior. We outline the algorithm below in detail.

First, we implement a GPU-accelerated culling technique to operate on our sparse transformed pointcloud, $\Psi'$. We start by projecting $\Psi'$ into 2D polar coordinates, 
\begin{align}
    \alpha = \arctan\left(\frac{\Psi'_y}{\Psi'_x}\right); \quad \beta = \arcsin\left(\frac{\Psi'_z}{d}\right); \quad d = \norm{\Psi'}_2
    \label{eq:lidarprojection}
\end{align}
where $(\alpha, \beta)$ are the yaw and pitch angles of the rays connecting each of the points, and $d$ are the distances along each ray. Now, the entire pointcloud, $\Psi'$ is represented as a sparse 2D image (without loss of information) over $(\alpha, \beta)$ with $d$ being the color or value of each pixel. To cull out points within our image, the distance of each pixel is compared to the average distance of its surrounding ``cone'' of neighboring rays. If the average distance of neighboring rays is less than the depth of the current pixel, the point is occluded and is removed from the sparse image. \figref{fig:lidar_culling} visualizes the large effect of our culling algorithm and the qualitative improvement it has on transformed pointclouds.

With our sparse and culled pointcloud, we need to build a dense representation of the scene to sample a new cast of LiDAR rays and generate the novel viewpoint. To densify our sparse representation we train a UNet architecture~\cite{ronneberger2015u} to learn a dense output of the scene. Training data for our densification network is generated using a 2D linear interpolator. We found that using a data-driven approach to densification yielded smoother, more natural qualitative results over strict rule-based interpolation ({\small\texttt{scipy.interpolate}}). Furthermore, the resulting model is easily GPU-parallelizable to achieve significant speedups ($\sim100\times$ faster). 

\begin{figure}[!t]
\centering
\includegraphics[width=\linewidth]{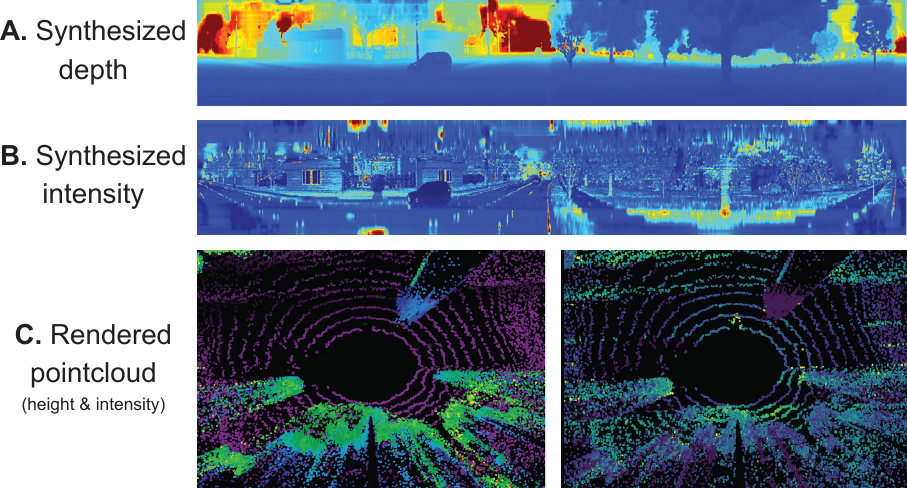}
\caption{\textbf{LiDAR novel view-synthesis.} Simulating a lateral translation of 1m off the road. Dense representations of depth (A) and intensity (B) are estimated from the sparse transformation. Sparse pointclouds (C) are rendered by sampling the dense representation according to the sensor prior.}
\label{fig:lidar_render}
\vspace{-10pt}
\end{figure}

Finally, we sample sparse points from our dense representation to form the novel view pointcloud. To determine sampling locations we we can construct a prior, $\Omega$, over the existing ray cast angles of the sensor in our dataset. The ray vectors for the sensor are largely fixed over time, as they are built into the hardware of the sensor, but can have some slight variations or drops based on the environment. Sampling $\omega$ from the prior yields a collection of rays, $\{(\alpha_i, \beta_i)\}$, to cast and collect point readings from. Furthermore, the prior, $\Omega$, will respect several desirable properties of the sensor which can also be user specified such as the quantity and density of the LiDAR rays. Since we are still operating in polar coordinate image space, $\omega$ is equivalent to a binary mask image denoting where in our dense image should be sampled. With our new, sampled polar image we can invert the transform in Eq.~\ref{eq:lidarprojection} to represent our data back in the desired 3D Cartesian space. \figref{fig:lidar_render} visualizes the different stages in the rendering pipeline, through the dense representation of the scene (A,B) as well as the result after sampling and reprojecting back to 3D cartesian space (C). 
\subsection{Event synthesis}

Event-based cameras are asynchronous, continuous-time sensors that detect brightness changes of the scene. An event is emitted when brightness change exceeds a certain threshold at a pixel location, and is described as a 4-tuple of pixel coordinate, timestamp, and polarity. The polarity is a binary value that indicates whether brightness change is positive or negative. Conceptually, event camera can be viewed as the derivative of regular RGB camera with additional advantages of much higher operating frequency ($> 10,000$Hz) and dynamic range. Given its similarity to RGB camera, event data can be simulated by taking derivative with respect to time over interpolated RGB frames \cite{rebecq2018esim,gehrig2020video}. Our proposed method extends prior work to additionally handle (1) non-aligned camera projection across RGB and event camera and (2) novel view synthesis according to vehicle's ego-motion. Simulating events from RGB instead of event data allows applying \vista~to existing datasets which mostly contain RGB sequences but not event data.

To capture the instantaneous change of pixel intensity, we need to first construct a continuous representation of RGB image stream. Given two consecutive RGB frames $I_{t_1},I_{t_2}$ and bidirectional optical flow $F_{t_1\rightarrow t_2},F_{t_2\rightarrow t_1}$, this can be achieved by arbitrary-time frame interpolation \cite{jiang2018super},
\begin{align}
    I_{t_1+k\Delta t} &= f_{\text{interp}}(I_{t_1},\, I_{t_2},\, F_{t_1+k\Delta t\rightarrow t_1},\, F_{t_1+k\Delta t\rightarrow t_2}) \\
    F_{t_1+k\Delta t\rightarrow t_1} &= -(1-k\Delta t) k \Delta t F_{t_1\rightarrow t_2} + (k \Delta t)^2 F_{t_2\rightarrow t_1} \\
    F_{t_1+k\Delta t\rightarrow t_2} &= (1-k\Delta t)^2F_{t_1\rightarrow t_2} - k\Delta t (1-k\Delta t)F_{t_2\rightarrow t_1}
    \label{eq:frame_interpolation}
\end{align}
where $k\in[0,\frac{t_2-t_1}{\Delta t}]$ and $k\Delta t$ specifies the time interval to be simulated in between. The arbitrary-time flow is derived from temporal consistency going forward ($t_1\rightarrow t_1+k\Delta t$) and backward ($t_2\rightarrow t_1+k\Delta_t$) in time with local smoothness assumption. The interpolation function $f_{\text{interp}}$ is implemented by a neural network that handles visibility issue from both directions. We refer the reader to \cite{jiang2018super} for more details. With this, we now apply an event generation model \cite{mueggler2017event,gallego2017event},
\begin{align}
    (\mathbf{p},t_k,\rho)\text{ if }\rho\big(\ln I(\mathbf{p},t_k)-\ln I(\mathbf{p},t_k-\Delta t)\big)\geq c_k
    \label{eq:event_generation_model}
\end{align}
where $\mathbf{p}$ is the pixel coordinate, $\rho\in\{-1,1\}$ is polarity and $c_k\sim\mathcal{N}(\mu_c,\sigma_c)$ is the contrast threshold sampled from a Gaussian to simulate noise. The temporal granularity of event generation is determined by $\Delta t$, which yields more accurate simulation with smaller values until saturating at subpixel displacement of optical flow. Furthermore, adaptive sampling \cite{rebecq2018esim} is used to jointly achieve accuracy and efficiency,
\begin{figure}[!t]
\centering
\vspace{-5pt}
\includegraphics[width=0.9\linewidth]{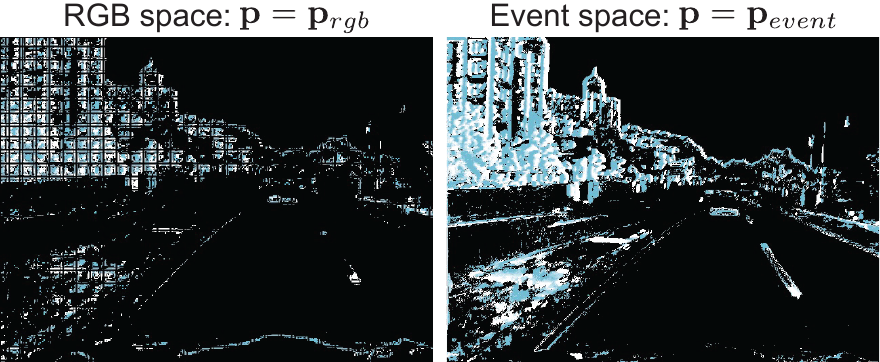}
\caption{\textbf{Different pixel spaces for event generation}. Events can be generated by estimating brightness change in RGB camera image space or virtual event camera image space.}
\label{fig:compare_event_sim}
\vspace{-10pt}
\end{figure}
\begin{align}
    \Delta t = \frac{t_2-t_1}{\min\{\underset{\mathbf{p}}{\max}\max\{F_{t_1\rightarrow t_2}(\mathbf{p}),F_{t_2\rightarrow t_1}(\mathbf{p})\}-1, \Delta t_{\max}\}}
    \label{eq:time_adaptive_sampling}
\end{align}
where $\Delta t_{\max}$ sets an upper bound to computational resources required for the simulation. Thus far, the only thing yet to be defined is the space that pixel coordinate $\mathbf{p}$ lives in. It can be image space of either RGB camera (input) or novel-view event camera (output), which are related by reprojection,
\begin{align}
    \mathbf{p}_{event}=K_{event}T_{event}^{novel}T_{rgb}^{event}D(\mathbf{p}_{rgb})K_{rgb}^{-1}\mathbf{p}_{rgb}
    \label{eq:pixel_transform}
\end{align}
where $K_{event},K_{rgb}$ are intrinsics for event and RGB cameras, $D$ is depth, and $T$ is transformation across two poses. We explicitly factorize $T_{rgb}^{event}$ since the control commands reference at existing sensors in the dataset. In event generation model \eqref{eq:event_generation_model}, using $\mathbf{p}_{rgb}$ involves generating events in RGB image space from the dataset and reprojecting pixel coordinates to event image space. We implement a bilinear sampler with thresholding to handle non-integer pixels after reprojection. On the other hand, using $\mathbf{p}_{event}$ renders the scene based on RGB dataset in event camera space and generate events without pixel reprojection. We argue that $\mathbf{p}=\mathbf{p}_{event}$ may be a better option since it casts the subpixel issue of reprojection \eqref{eq:pixel_transform} from interpolating in pixel coordinate space as in $\mathbf{p}=\mathbf{p}_{rgb}$ to interpolating in color/intensity space of meshes during RGB image rendering. The comparison can be seen in \figref{fig:compare_event_sim}.





\section{Policy learning}
\subsection{Data generation and training using VISTA}
\label{sec:policy_learning}
The capability of \vista~to simulate sensor measurements allows for reinforcement learning~\cite{amini2020learning} as well as guided policy learning by generation of novel view training data. With access to the internal state of the simulator, a controller with privileged information \cite{levine2013guided,bojarski2016end,chen2020learning} (\GPLplain) can generate optimal control commands corresponding to agent's current state. Associating this control with perception data increases diversity of training data distribution and allows generating edge cases locally around the dataset, thus improving the recovery robustness of the policy. Algorithm~\ref{alg:gpl} describes data generation in \vista~and \GPLplain. First, the simulator is reset with random initialization (e.g., off-center position and heading). The simulator is stepped with control commands from the privileged controller and iteratively generates sensor measurements and optimal control labels for supervising policy learning. While there are no restrictions on the optimal controller used for the privileged agent, we use a pure-pursuit controller in our experiments. We use a shuffled buffer to approximately ensure \textit{i.i.d.} training samples. Before adding data into the buffer,  rejection sampling is used to balance the label distributions~\cite{amini2019variational}. Finally, we introduce a branching step that locally branches out from stepping the simulator with privileged control (e.g., turning right instead of left as instructed). This is extremely important to policy learning with event cameras since event data captures \textit{changes in the scene} and thus conflates the vehicle's ego-motion with its own control. This means that a policy can accurately correlate control to the motion of the scene instead of attending to the road. While not presenting an obvious issue during open-loop evaluation~\cite{maqueda2018event}, these policies will fail catastrophically on a closed-loop test. However, by branching with arbitrary control, we effectively disentangle ego-motion and scene information in event patterns.
\begin{algorithm}
    \footnotesize 
    \caption{Data generation and training in VISTA}
    \label{alg:gpl}
    \begin{algorithmic}
        \For{$k \gets 1$ to $N$}
            \While{! \texttt{buffer.full()}}
                \If{\texttt{VISTA.done()}}
                    \State \texttt{VISTA.reset()}
                \EndIf
                \State $x \gets \texttt{VISTA.readSensorBuffer()}$
                \If{\texttt{useBranching}}
                    \State \texttt{VISTA.randomStep()}
                    \State $y \gets \texttt{privilegedController(\texttt{VISTA.getState()})}$
                    \State \texttt{revertState(VISTA, privilegedController)}
                \Else
                    \State $y \gets \texttt{privilegedController(\texttt{VISTA.getState()})}$
                \EndIf
                \State \texttt{VISTA.step($y$)}
                \If{! \texttt{rejectSample($x, y$)}}
                    \State \texttt{buffer.add($x, y$)}
                \EndIf
            \EndWhile
            \State \texttt{buffer.shuffle()}
            \State \texttt{trainModel(buffer.next())}
        \EndFor
    \end{algorithmic}
\end{algorithm}
\vspace{-10pt}
%
\subsection{Input representation and model architectures}
All models consist of a feature extractor which processes the sensory data and a deterministic estimator which learns control from the features. We use the same network architecture for the estimator (a 3-layer fully connected network), which outputs a scalar value as curvature. For RGB images, we use a simple 5-layer CNN, each layer comprises a convolution, group norm, and ReLU~\cite{amini2019variational}. For event data, we accumulate events within a small time interval, project them on a frame according to their pixel coordinates and apply a convolutional feature extractor~\cite{maqueda2018event}. For LiDAR data, we use FastLiDARNet \cite{liu2021efficient,tang2020searching} that can process point cloud efficiently with sparse tensor operation. 


\section{Results}

\subsection{Experimental setup and data collection}
\noindent\textbf{Hardware setup.} We collect data and deploy learned policies on a full-scale vehicle (2019 Lexus RX 450H) which we have outfitted with autonomous driving capabilities. The car is equipped with an NVIDIA 2080Ti GPU and an AMD Ryzen 7 3800X 8-Core Processor. Perception sensors include a 30Hz BFS-PGE-23S3C-CS RGB camera 
, a 10Hz Velodyne VLS-128 LiDAR sensor, and a Prophesee Gen3 event-based camera. The event camera runs at adaptive rate based on events emission which range from hundreds to thousands Hz. Other on-board sensors include inertial measurement units (IMUs) and wheel encoders for estimating odometry as well as a centimeter-level accurate OxTS global positioning system (d-GPS) for evaluation.

\noindent\textbf{Data collection.} We collect data from multiple sensors (RGB camera, LiDAR, event camera) in a wide variety of environments, including different time of day (daytime/night), weather conditions (sun/rain), and road types (urban/rural). The entire dataset contains roughly 3 hours of driving data. RGB images, LiDAR point cloud, event data, and curvature feedback are used for \vista~simulation, policy learning and evaluation. GPS data is only used for evaluation.
%
%
%
\begin{table}[b!]
\centering
\begin{tabular}{@{}cc|cccccc@{}}
\toprule
\multirow{2}{*}{\textbf{Sensor}} & \multirow{2}{*}{\textbf{Algo.}} & \multicolumn{6}{c}{\textbf{Mean Squared Error (1E-5)}} \\
& & Day & Night & Sun & Rain & Urban & Rural \\
\midrule
\multirow{2}{*}{RGB} & \ILplain & 0.64 & 0.15 & 0.51 & 0.39 & 0.69 & 2.10 \\ 
& \GPLplain & 3.01 & 0.61 & 22.44 & 2.34 & 14.92 & 4.75 \\
\midrule
\multirow{2}{*}{LiDAR} & \ILplain & 0.92 & 1.04 & 0.74 & 2.52 & 1.02 & 0.61 \\ 
& \GPLplain & 7.35 & 9.83 & 8.28 & 11.12 & 8.87 & 5.07 \\
\midrule
\multirow{2}{*}{Event} & \ILplain & 0.18 & 0.26 & 0.37 & 1.24 & 0.19 & 10.81 \\ 
& \GPLplain & 20.75 & 10.12 & 16.54 & 6.54 & 10.88 & 10.25 \\
\bottomrule
\end{tabular}
\caption{\textbf{Open-loop control errors.} Open-loop IL outperforms GPL when considering only error. Note that low open-loop error is a very poor indicator for closed-loop success~\cite{codevilla2018offline}.}
\label{tab:open_loop_error}
\end{table}
\begin{table*}[h!]
\centering
\begin{tabular}{@{}cc|cccccc|cccccc@{}}
\toprule
\multirow{2}{*}{\textbf{Sensor}} & \multirow{2}{*}{\textbf{Algo.}} & \multicolumn{6}{c}{\textbf{Mean Deviation}} & \multicolumn{6}{c}{\textbf{Crash Rate}} \\ 
& & Day & Night & Sun & Rain & Urban & Rural & Day & Night & Sun & Rain & Urban & Rural \\
\midrule
\multirow{2}{*}{RGB} & \ILplain & 0.283 & 0.165 & 0.289 & 0.425 & 0.285 & 0.191 & 0.080 & 0.008 & 0.090 & 0.010 & 0.146 & 0.094 \\ 
& \GPLplain & 0.102 & 0.068 & 0.120 & 0.101 & 0.120 & 0.226 & 0.002 & 0.000 & 0.002 & 0.006 & 0.004 & 0.000\\
\midrule
\multirow{2}{*}{LiDAR} & \ILplain & 0.327 & 0.302 & 0.295 & 0.366 & 0.307 & 0.213 & 0.664 & 0.656 & 0.652 & 0.734 & 0.668 & 0.330 \\ 
& \GPLplain & 0.266 & 0.258 & 0.280 & 0.323 & 0.268 & 0.274 & 0.334 & 0.330 & 0.322 & 0.426 & 0.316 & 0.150 \\
\midrule
\multirow{2}{*}{Event} & \ILplain & 0.340 & 0.344 & 0.327 & 0.320 & 0.329 & 0.362 & 0.486 & 0.828 & 0.714 & 0.576 & 0.674 & 0.784 \\ 
& \GPLplain & 0.307 & 0.313 & 0.293 & 0.278 & 0.319 & 0.376 & 0.166 & 0.200 & 0.084 & 0.051 & 0.324 & 0.442 \\
\bottomrule
\end{tabular}
\caption{\textbf{Closed-loop performance of policies in \vista.} GPL policies exhibit consistently reduced deviation from the center line than compared to real-world IL. Furthermore, these polices exhibit greater robustness with lower crash rates.}
\label{tab:closed_loop_vista}
\end{table*}
\begin{figure*}[!t]
\centering
\includegraphics[width=0.95\linewidth]{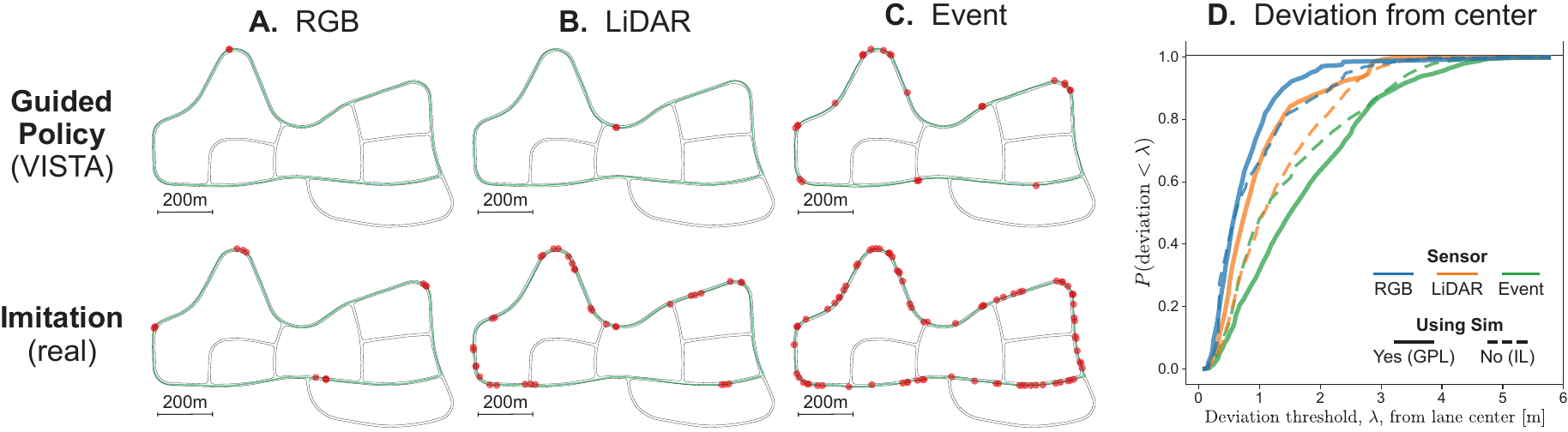}
\caption{\textbf{Real-world deployment.} Policy trajectories (n=3; 45km total) and crash locations (red dots) for RGB (A), LiDAR (B), and Event (C) sensors. Cumulative distribution of deviations from the center (D) shows the benefit of training in \vista. }
\label{fig:devens_loops}
\end{figure*}
\begin{table}[b!]
\centering
\begin{tabular}{@{}c|ccc@{}}
\toprule
\textbf{Recovery Rate} & \textbf{RGB} & \textbf{LiDAR} & \textbf{Event} \\
\midrule
\ILplain & $0.27\pm0.13$ & $0.16\pm0.11$ & $0.00 \pm 0.00$ \\
\GPLplain & $\bm{0.90\pm0.13}$ & $\bm{0.83\pm0.22}$ & $\bm{0.90\pm0.15}$ \\
\bottomrule
\end{tabular}
\caption{\textbf{Robustness Test.} \GPLplain~policies significantly outperform real-world \ILplain~at recovering from edge cases.}
\label{tab:robustness_test}
\end{table}

\noindent\textbf{Evaluation metrics.} For open-loop tests, we compute mean squared errors between human control command (curvature) and policies' predictions. For closed-loop tests, we compute mean deviation from the lane center, crash rate, and number of interventions. In simulation, we consider crashes as lateral translation from human trajectories larger than 2 meters. In real-world tests, we intervene and take over control from autonomous mode once vehicle is off the road.

\subsection{Offline evaluation}
In \tabref{tab:open_loop_error}, we show open-loop control errors of \ILplain~and \GPLplain. Note that the error is of the scale $1e^{-5}$ (normal driving roughly has curvature between $\pm0.05$) and thus both algorithms perform quite well in spite of the difference. \ILplain~outperforms \GPLplain~as expected given \ILplain's training objective is aligned with such evaluation. However, open-loop error is a very poor indicator of measuring the performance of a driving policy~\cite{codevilla2018offline} since it (1) only measures errors around human trajectories and (2) ignores compounding feedback errors that bring the vehicle to out-of-distribution states. 

Instead, closed-loop test settings (either in simulation or reality) serve as a greater proxy for evaluating policy performance. We start by using \vista~ to measure closed-loop performance with synthetic data before moving to the real-world. \tabref{tab:closed_loop_vista} shows closed-loop performance of policies in \vista. Consistent with prior research, we qualitatively observe a common failure mode of \ILplain~ policies where they that gradually drift off the road and cannot recover (due to lack of recovery training data). \GPLplain~ policies exhibit consistently reduced deviation from the lane center and lower crash rate than \ILplain~ policies in a wide range of environments, lighting, and sensors modalities.

\subsection{Online real-world test}
In \figref{fig:devens_loops}, we demonstrate real-world policy deployment of \ILplain~ and \GPLplain~ policies. For each policy, we run the vehicle autonomously (controlled by the policy) for 3 trials in the outerloop of the test track (total distance of all trials is 45km). \figref{fig:devens_loops}(A-C) show interventions (red dots) throughout multiple trials. \figref{fig:devens_loops}(D) shows percentage of deviation from center smaller than a range of thresholds, where larger area below the line means more stable lane keeping maneuvers. The performance of \GPLplain~policies for RGB and LiDAR are significantly better than \ILplain~policies. This is highly aligned with our observation from closed-loop testing in \vista (\tabref{tab:closed_loop_vista}), which further motivates its effectiveness for policy evaluation, considering the time and safety costs of real-world testing.
For event camera, while \GPLplain~ policy also exhibit superior performance in terms of number of interventions, it deviates more from the center compared to \ILplain~policy and suffers from much frequent intervention compared to RGB and LiDAR \GPLplain~policies. This is due to the fact that event cameras can only see the component of the road boundary non-parallel to vehicle's ego motion, while RGB and LiDAR sensors provide sufficient information at every step for lane following. Such properties highlight the potential utility of fusing event sensing with RGB and LiDAR for greater benefits. We observed a swirling maneuver along straight roads with event policies (8x more jittery than RGB (measured by squared second derivative of curvature). However, we found that our method of branched learning for preventing ego-motion conflating the scene understanding (\secref{sec:policy_learning}) is highly effective in real-world tests to reduce the number of interventions: $28.0\pm3.6$ (\ILplain) $18.0\pm1.0$ (\GPLplain) $6.5\pm3.8$ (\GPLplain~with branching).
To further highlight the effectiveness of \GPLplain~policies, we conduct a robustness test by initializing the car with $\pm 30^\circ$ rotation and $\pm 2m$ translation from the lane center and measure the success rate of recovery, as shown in \tabref{tab:robustness_test}.




\section{Conclusion}
We present \vista, an open-source simulator that supports multimodal sensor synthesis including 2D RGB cameras, 3D LiDAR, and event-based cameras for mobile agents. The simulator is data-driven and capable of synthesizing high-fidelity sensor measurement sufficient for policy learning and evaluation. We showcase the sim-to-real ability by directly deploying policies learned in \vista~on a full-scale autonomous vehicle for each sensor and demonstrate consistent results between closed-loop evaluation in simulation and real-world test. We believe the release and scalability of \vista~ opens up new research opportunities to the community for perception and control of autonomous vehicles.



\bibliographystyle{IEEEtran}
\bibliography{ref}

\end{document}